\documentclass[conference]{IEEEtran}
\IEEEoverridecommandlockouts

\newif\ifanon
\anonfalse

\usepackage[T1]{fontenc}
\usepackage{amsmath,amssymb}
\usepackage{graphicx}
\usepackage{booktabs}
\usepackage{listings}
\usepackage{upquote}
\usepackage{url}
\usepackage[hidelinks]{hyperref}

\begin{document}

\title{Empirical Evaluation of Task-Based Permission\\Scoping Architecture for AI Agents}

\ifanon
\else
  \author{\IEEEauthorblockN{Halil Burak Noyan}
  \IEEEauthorblockA{\textit{Independent Researcher}\\
  contact@buraknoyan.com}}
\fi

\maketitle

\ifanon\else
  \pagestyle{plain}\thispagestyle{plain}
\fi

\begin{abstract}
AI agents are provisioned the same as employee-owned hosts in many enterprise settings with a static credential set fixed at deployment which includes all permissions the employee role might ever need. Role-based access control made this compromise for human principals because scoping access per task was infeasible. For AI agents, the compromise leaves every credential standing exposed whether or not the current task uses them. These permissions can later be utilised by a compromised or misaligned agent. Prior work (Noyan, 2026) defined this as the task-context mismatch, and proposed a three-source permission architecture which includes role-based permission ceilings, a task permission classifier and policy-based prohibitions, together eliminating the exposure preemptively. The work released a 600-prompt labelled dataset to evaluate it.

This paper presents that evaluation end to end by implementing the security gate; a fine-tuned RoBERTa-large encoder which matched few-shot trained Claude Haiku 4.5 on classification quality (macro-F1 0.881 against 0.886, precision 0.897 against 0.842, severity-weighted residual risk 0.63 against 1.12). The results show the trusted component does not need to scale with the agent it supervises, and the scalable-oversight margin for this control method is wide.

We also propose an attack-surface elimination metric which shows the role ceiling alone closes 27.9\% of the severity-weighted surface and adding the task classifier closes 84.4\%. The gap displays security advantages of task-granular access control over role-granular, and AI agents are the first principal type for which the task-granular access control is enforceable because their tasks arrive as machine-readable text.

The prohibition layer doesn't add measurable benefit on top of the deployed classifier. When the classifier is bypassed entirely, it cuts 11.5-13.6\% of three departments' credential ceilings while blocking 6-31\% of those departments' legitimate tasks. The experiments show that combination prohibitions must be enforced at runtime with fine-grained permission taxonomy, and they are effective once other layers are ablated.

The research establishes task-based access control as a measured, potentially deployable mechanism for reducing attack surface in agentic deployments.
\end{abstract}

\section{Introduction}
Enterprise AI agents are typically provisioned with a static credential envelope covering everything their role might ever need. This results in a task-context mismatch (Noyan, 2026). A credential absent from an agent's context cannot be misused regardless of the agent's reasoning or evasion sophistication. Therefore we treat capability scoping as prevention rather than detection against agents that adapt to oversight (Greenblatt et al., 2024; Meinke et al., 2024). The prior work proposed a concrete three-source architecture instantiating this principle and released the dataset needed to evaluate it, deferring the evaluation itself to future work. This paper presents that evaluation.

The paper makes four contributions.

\begin{itemize}
  \item An implementation and evaluation of the task-context classifier across three architectures, establishing that a 355M-parameter on-host encoder reaches a frontier model's classification quality (Section 4).
  \item The severity-weighted attack surface elimination metric that shows the full system closes 84.4\% of the exposure created by static provisioning (Section 6).
  \item The design, implementation, and evaluation of the policy-derived prohibition layer that provides 11.5-13.6\% attack surface reduction when the classifier is bypassed (Sections 5 and 6).
  \item A methodological result that shows task-based access control is potentially deployable in enterprise settings and in serverless infrastructure (Section 7).
\end{itemize}
Section 2 situates the work in prior literature. Section 3 fixes the setup shared by the rest of the paper. Section 4 evaluates the task-context classifier against two baselines, Section 5 derives the prohibition rule set, and Section 6 evaluates the assembled architecture end-to-end. Section 7 discusses the results in depth, Section 8 outlines future work, and Section 9 concludes.

\section{Related Work}
Capability-based security predates LLM agents by over sixty years (Dennis \& Van Horn, 1966), with least privilege formalised by Saltzer \& Schroeder (1975) and realised in systems such as EROS (Shapiro et al., 1999) and CapDesk (Miller, 2006). Classical capability systems trust the principal to hold its own credentials and attenuate them when delegating. An LLM agent must not be trusted this way, so the scoping decision has to sit outside the agent, in the orchestrator and credential service that is marked as trusted. Zero trust architecture (Rose et al., 2020) proposes a deployment method treating every entity as untrusted by default regardless of perimeter, but doesn't provide a threat model for AI agents. OWASP's LLM06, Excessive Agency (OWASP Foundation, 2025) covers this gap by naming persistent over-privilege as a defining failure class for deployed agents.

Guardrail systems such as Llama Guard (Inan et al., 2023) and NeMo Guardrails (Rebedea et al., 2023) enforce policy by filtering inputs and outputs at the agent boundary, and behavioural monitors intervene on runtime. These detection mechanisms have fundamental limitations against agents that reason about oversight and adapt to it (Greenblatt et al., 2024; Meinke et al., 2024). The attacker (who is an AI agent or an operator in this case) needs only one phrasing the monitor misses, and indirect prompt injection supplies a steady stream of candidate phrasings (Greshake et al., 2023). The architecture proposed by this paper acts before the detection layer by removing the credential the attack would need. The observe-only mode defined in Section 3.1 produces the behavioural signal which feeds into the detection layer so the two approaches complement each other.

PCAS (Palumbo et al., 2026) compiles a formal Datalog policy specification into an authorisation layer, the role the prohibition layer (Source 3) plays here. The difference is that in PCAS someone must write and maintain a second artefact as the specification, but in our proposal the rules are self-maintained and kept up-to-date since they derive from the organisation's natural-language policy document in the spirit of Policy-as-Prompt (Kholkar \& Ahuja, 2025). Conseca (Tsai \& Bagdasarian, 2025) generates just-in-time policies each time the agent tries an action, and evaluates the action against them, whereas the task-context classifier predicts the full permission set up front from the task description which cannot be influenced by a compromised agent at runtime.

In the AI control framing, a trusted, weaker system supervises an untrusted, stronger one, and the capability gap between the two is the central difficulty (Greenblatt et al., 2024). Sleeper agents (Hubinger et al., 2024) and in-context scheming (Meinke et al., 2024) motivate control protocols that do not rely on the supervised model's cooperation. Section 7 argues that the task-context classifier is not limited by this capability margin. It also discusses how OpenClaw evaluations (Wang et al., 2026) document what goes wrong when agent hosts hold standing credential sets, and how per-task scoping removes that pattern.

Existing enterprise tools already limit long-lived credentials. Just-in-time privileged access grants time-boxed credentials on request with human or workflow approval, and cloud workload identity attaches ceilings to service principals exactly as the role ceiling does (Amazon Web Services, 2024). Both operate at the granularity of a request session or a principal identity, and neither derives the permission set from the content of the task. The classifier evaluated in Section 4 adds that derivation, which is the step the role-versus-task argument of Section 7 turns on.

\section{Background and Experimental Setup}
\subsection{Architecture, Threat Model, and Deployment Modes}
\begin{figure}[t]
  \centering
  \includegraphics[width=\columnwidth]{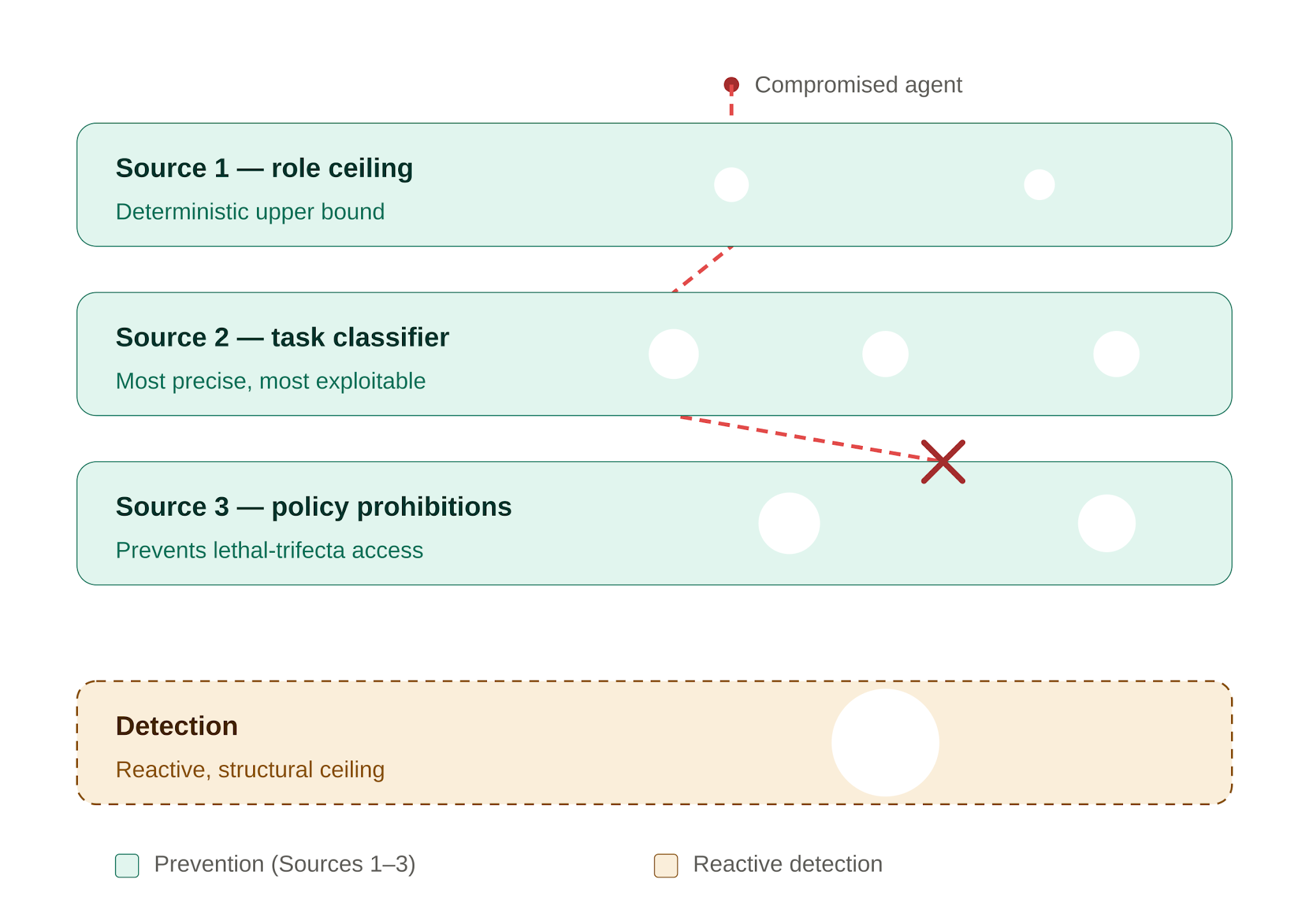}
  \caption{Layers of the three-source architecture presented sequentially: role-based permission ceilings, task permission classifier and policy-based prohibitions.}
  \label{fig:architecture}
\end{figure}

The architecture under evaluation grants a permission only if three independent sources agree: Source 1, a deterministic per-department role ceiling; Source 2, a classifier looking at the task description and predicting the required permission set; and Source 3, deterministic prohibitions of permission combinations derived from the organisation's policy. Each source covers a different failure mode. Ceilings cannot distinguish tasks within a role, the classifier operates on manipulable natural language, and prohibitions are deterministic but coarse (Noyan, 2026, §3). They form a security-in-depth architecture together. The paper refers to them below as the role ceiling, the task-context classifier, and the prohibition layer.
Figure~\ref{fig:architecture} shows the arrangement.

The actor in the threat model is a compromised or misaligned agent operating inside the permission set it has been granted. The classifier, the credential deployment service, and the orchestrator are trusted components. Trusting an LLM-based classifier as the security gate remains the strongest of these assumptions. The classifier operates on a natural-language task description that an attacker can influence through prompt injection or context pollution, and it will faithfully over-grant on a poisoned description. The evaluation also covers scenarios where the classifier is bypassed and the system falls back to the role ceiling and policy prohibitions. An attacker can wait for a task issued by a highly privileged user to get the highest privileges, but the prohibition rules apply to the permission combination requested regardless of who is asking. Bypassing other layers through privilege escalation exploits or side channel attacks in the runtime is possible and discussed in Section 6.5.

The architecture has two independent deployment choices (Noyan, 2026). An enforcing deployment blocks the task, an observe-only deployment logs it as an artefact for further examination. A static deployment checks the complete candidate permission set before a task begins which makes it a preemptive solution. A runtime deployment monitors the sequence of permissions real-time during execution. All quantitative results in this paper are produced under the static mode; Section 6.5 analyses what the runtime mode changes.

\subsection{Dataset and Metrics}
The evaluation dataset contains 600 enterprise task prompts grounded in a six-department synthetic company policy (TechCorp), each labelled with the minimum required permissions from a 15-permission tool-based taxonomy. Every permission in the taxonomy maps to a deployable credential type and carries one of three risk tiers (Tier 1: highest risk, Tier 3: lowest risk). Prior work validated the labels against independent expert re-annotation at Cohen's $\kappa$ (Cohen, 1960) = 0.967 (Noyan, 2026, §5). The dataset release includes the policy document, company\_policy.txt which is the source of the dataset's ground-truth labels and of the prohibition rules derived in Section 5. The classifier is trained on the 500-record training split and evaluated on the 100-record held-out test split. Classifier-dependent metrics exist only for the test split, while properties of the role ceilings and the rule set are checked against all 600 records since they do not depend on a trained model.

Every severity-weighted metric weights permissions by risk tiers explained above, 3/2/1 for Tier 1/2/3. The taxonomy holds four Tier 1, eight Tier 2, and three Tier 3 permissions, so granting them all means a severity-weighted risk score of 31.

Overshoot metric measures the fraction of test records granted at least one permission beyond ground truth, and undershoot measures the fraction denied at least one required permission. Auto-handled metric is the complement of undershoot and shows what percentage of tasks complete without an under-grant forcing a retry or manual escalation.

Severity-weighted delta sums the tier weight of every over-granted permission on a record, averaged across the test set. It measures the residual risk in a classifier's output. A system granting almost nothing and a system granting everything can each report a small or large severity-weighted delta depending on the criticality of the over-granted permissions.

Attack surface elimination metric shows how much risk reduction is obtained by each mechanism calculating the severity-weighted sum of the removed permissions. This measurement is compared with the grant-everything baseline and it is reported as a percentage. For example, a mechanism that eliminates 85\% of a baseline's weighted surface has effectively closed 85\% of the cyber security risk that would otherwise be exposed. Elimination counts every permission removed from the baseline including the faulty restrictions. It must be read jointly with undershoot to measure true performance. Section 6 uses elimination to evaluate the risk reduction of the whole system.

\subsection{Threshold Configurations and the Deployability Standard}
Each permission is associated with a risk level (Tier 1, default deny, up to Tier 3, default allow). The threshold setup assigns a confidence threshold for each risk level rather than all 15 permissions. A Tier 1 permission needs higher model confidence to be granted than a Tier 3 permission. Six configurations were evaluated: two flat cutoffs applied to all fifteen permissions (0.5 and 0.8) and four risk-tiered Tier 1/2/3 cutoffs (0.7/0.5/0.3, 0.6/0.4/0.2, 0.5/0.3/0.1, 0.8/0.6/0.4). The canonical configuration (0.7/0.5/0.3) is used for the head-to-head comparison in Section 4.2, and the recommended configuration (0.8/0.6/0.4) is the one used in the prohibition-layer evaluations of Section 6.

Before conducting the experiments, we have committed to a deployability standard which states that the undershoot has to be less than 10\%, combined with the macro precision no lower than 0.90 for the solution to be realistically deployed in enterprise settings in enforcement mode. Section 4.3 judges the swept configurations against the standard.

\section{Evaluating the Task-Context Classifier (Source 2)}
\subsection{Candidate Classifiers}
For evaluation of task-context classifiers, the paper implements a fine-tuned RoBERTa-large encoder model. The model gets compared to a weak baseline (TF-IDF with logistic regression) which establishes a floor. Then it is compared to an off-the-shelf frontier model (Claude Haiku 4.5) using few-shot learning to explore whether a fine-tuned, self-hosted encoder can match a much larger general-purpose model.

\textbf{TF-IDF baseline.} Term-frequency features over unigrams and bigrams (max\_features=10000) feed a one-vs-rest logistic regression. It has been swept across 6 configurations (Flat 0.5, Flat 0.8, and 4 severity-weighted thresholds) and the 0.5 decision threshold is chosen as baseline which has the best macro-precision (0.765) and macro-F1 (0.702) scores. This baseline exists to confirm that a specialised model, fine-tuned or otherwise, is doing load-bearing work on this task rather than solving a problem a classical method already solves.

\textbf{Claude Haiku 4.5, few-shot.} claude-haiku-4-5-20251001 at temperature 0, given a system prompt containing the full 15-permission taxonomy and 6 hand-picked examples drawn from the training split. The prompt asks for a confidence value per permission so they can be swept with different threshold configurations reported below.

\textbf{RoBERTa-large, fine-tuned.} A pretrained roberta-large checkpoint (355M parameters) fine-tuned as a multi-label sequence classifier, one sigmoid output per permission. Training used the 500-record training split for 40 epochs, batch size 16, learning rate 2e-5, weight decay 0.01, warmup ratio 0.1, maximum sequence length 256 tokens, seed 42, fp32 precision. The loss is binary cross-entropy with per-class positive weighting (pos\_weight, clipped at 10) to counteract the taxonomy's class imbalance, most permissions are denied for most tasks, and gradients are clipped at norm 1.0. Inference on the 100-record test set is cached as a probability matrix and swept with six threshold configurations.

An identically configured RoBERTa-base was also run which was considerably worse than RoBERTa-large on macro-precision and macro-F1 in every configuration. Therefore it is not included in the comparison.

\subsection{Classification Performance}

Table 1 reports the three candidate classifiers at their canonical settings.

\begin{table}[t]
\caption{Test-set performance of the three candidate classifiers.}
\label{tab:1}\centering\footnotesize\setlength{\tabcolsep}{5pt}
\begin{tabular}{lccc}
\toprule
\textbf{Metric} & \textbf{TF-IDF} & \textbf{Claude} & \textbf{RoBERTa-large} \\
 & & \textbf{Haiku 4.5} & \textbf{(0.7/0.5/0.3)} \\
\midrule
Severity-weighted delta & 0.95 & 1.12 & 0.63 \\
Overshoot & 33\% & 42\% & 23\% \\
Undershoot & 43\% & 11\% & 21\% \\
Macro-precision & 0.765 & 0.842 & 0.897 \\
Macro-F1 & 0.70 & 0.886 & 0.881 \\
\bottomrule
\end{tabular}
\end{table}

The baseline confirms the fine-tune is load-bearing. TF-IDF trails RoBERTa-large on every metric at its best operating point. Even RoBERTa-large's loosest configuration (0.5/0.3/0.1, Table 2) stays ahead of TF-IDF on macro-precision and macro-F1. Against Claude Haiku, TF-IDF trails on classification quality and availability (macro-F1 0.702 and undershoot 43\% against macro-F1 0.886 and undershoot 11\%). Its lower severity-weighted delta and overshoot are due to its bias to grant little rather than being precise, as shown by 43\% undershoot rate.

Claude Haiku and RoBERTa-large, at their respective canonical settings, trade off in opposite directions. Haiku's higher recall gives it lower undershoot and a higher auto-handled rate. RoBERTa-large's higher precision gives it a lower severity-weighted delta and lower overshoot. Macro-F1 averages the two and ties them within half a point (0.886 against 0.881).

The two failure modes of Claude Haiku and RoBERTa-large have separate tradeoffs. An over-grant expands the credential surface available to a compromised or manipulated agent whereas an under-grant degrades task completion but grants nothing an attacker could use. A permission classifier intended as a security gate should therefore be tuned to deny high-risk permissions and permit low-risk ones, a precision-favouring bias. Macro-precision (0.897 against 0.842) and severity-weighted delta (0.63 against 1.12) are the precision-side metrics, and RoBERTa-large leads on both.

From an availability standpoint, RoBERTa-large has a clear disadvantage as 21\% of tasks had at least one required permission stripped away, compared to 11\% with few-shot Claude Haiku. This means the tasks with unsatisfiable permissions either prompt the user for the extra grants, or simply fail in an enterprise deployment setting.

\begin{figure}[t]
  \centering
  \includegraphics[width=\columnwidth]{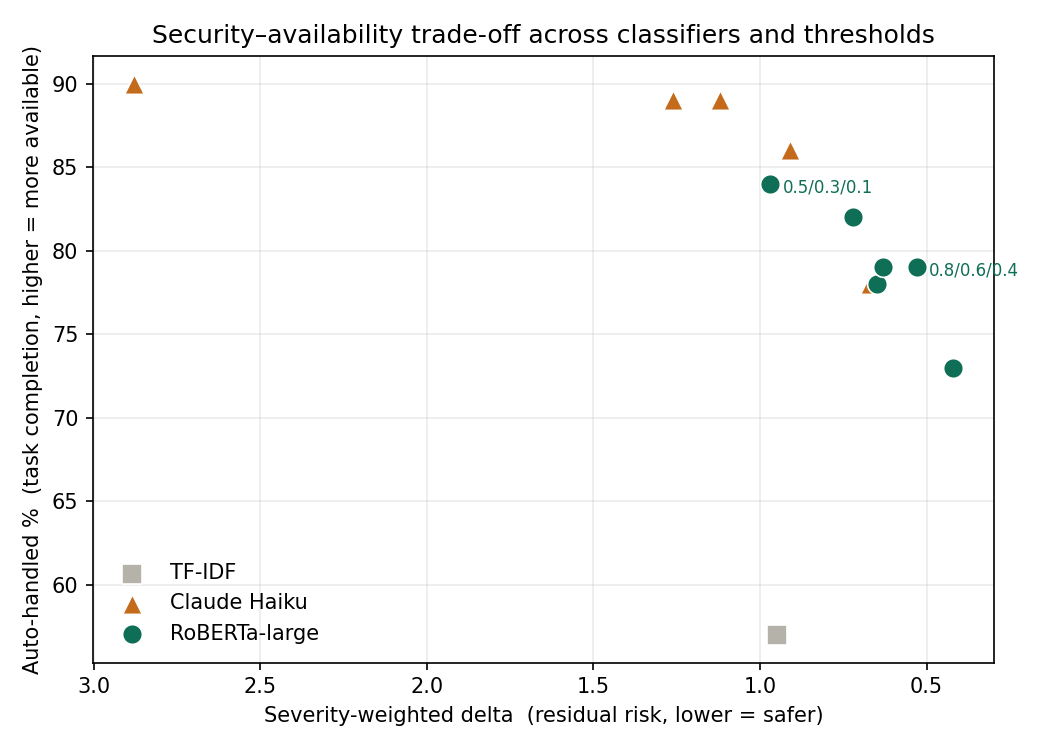}
  \caption{Security-availability trade-off across classifiers and threshold
  configurations.}
  \label{fig:tradeoff}
\end{figure}

Table 2 contains the statistics from full sweep on RoBERTa-large and Claude Haiku models, then compares severity-weighted delta (residual risk) and auto-handle rates (task completion) for each configuration. Figure~\ref{fig:tradeoff} plots that surface.

\begin{table*}[t]
\caption{Six configuration threshold sweep on target classifiers.}
\label{tab:2}
\centering
\footnotesize
\setlength{\tabcolsep}{4pt}
\begin{tabular}{lcccccc}
\toprule
\textbf{Configuration} & \textbf{RoBERTa delta} & \textbf{RoBERTa auto-handled} & \textbf{RoBERTa precision} & \textbf{Haiku delta} & \textbf{Haiku auto-handled} & \textbf{Haiku precision} \\
\midrule
Flat 0.5 & 0.65 & 78\% & 0.898 & 1.12 & 89\% & 0.848 \\
Flat 0.8 & 0.42 & 73\% & 0.941 & 0.67 & 78\% & 0.895 \\
0.7/0.5/0.3 & 0.63 & 79\% & 0.897 & 1.12 & 89\% & 0.842 \\
0.6/0.4/0.2 & 0.72 & 82\% & 0.885 & 1.26 & 89\% & 0.816 \\
0.5/0.3/0.1 & 0.97 & 84\% & 0.832 & 2.88 & 90\% & 0.697 \\
0.8/0.6/0.4 & 0.53 & 79\% & 0.915 & 0.91 & 86\% & 0.865 \\
\bottomrule
\end{tabular}
\end{table*}

RoBERTa-large dominates the safety half of this surface. At its safest configuration, the flat 0.8 cutoff, it reaches severity-weighted delta 0.42, a residual-risk level Claude Haiku does not attain at any threshold, since Haiku's safest point is 0.67, also at flat 0.8. For any Haiku operating point below auto-handled 84\%, a RoBERTa-large configuration exists that is at least as available at lower residual risk. Claude Haiku wins only the high-availability corner above auto-handled 84\%, reaching 86-90\% where RoBERTa-large tops out at 84\%. This availability comes with the increased risk as Haiku's 89\% configurations sit at severity-weighted delta 1.12 or above, against 0.53 at RoBERTa-large's recommended configuration and 0.97 at its most permissive.

The frontier maps onto the enforcement axis defined in Section 3.1. In observe-only deployment RoBERTa-large's precision advantage carries no availability cost, therefore it is the preferred classifier. In enforcing deployment its 21\% undershoot rate means these tasks would fail without completion. In environments with high availability requirements, Claude Haiku's auto-handle rate outweighs the added residual risk.

The fine-tuned classifier's residual over-granting is not spread evenly across the taxonomy. At the recommended configuration, three of the fifteen permissions, database\_read, email\_read, and confluence\_read, account for 58\% of all over-grant events, and database\_read alone accounts for 25\%. When each over-grant is weighted by their risk tier, the top two permissions (database\_read, email\_send\_external) produce 51\% of the total residual risk. This is because database\_read is a Tier 1 permission and its over-grants count triple, producing 34\% of the severity-weighted residual overshoot alone.

Certain underrepresented permissions in the dataset cause the most classification failures. We have evaluated a hypothetical hybrid architecture which routes email\_send\_external through a deterministic allowlist as a proof of concept. It raised macro-precision from 0.915 to 0.940 and macro-F1 from 0.890 to 0.915 while cutting undershoot from 21\% to 18\%.

In contrast, undershoot does not concentrate. It is spread near-uniformly across twelve permissions and there is no deterministic patch that can reduce it enough to meet our self-committed deployability standard which was explained in Section 3.3.

\subsection{Deployment Considerations}
\textbf{Compute time and cost.} Although RoBERTa-large runs in the network perimeter and Claude Haiku is an external API call, we can assume the network round-trip times are identical. Published benchmarks put time-to-first-token for Claude Haiku 4.5 in the range of 0.6 to 1.0 seconds depending on provider and prompt length. Published CPU inference benchmarks for encoder models in the same class as RoBERTa-large vary widely by hardware and input length. They tend to be around tens of milliseconds on optimised short-sequence setups and several hundred milliseconds on longer inputs without optimisation. This paper did not benchmark the two solutions side by side on identical hardware, and the direct comparison is deferred to a controlled measurement. However, RoBERTa-large is expected to have a significant advantage in response time.

\textbf{Calibration and drift.} Claude Haiku's behaviour at a given threshold configuration is calibrated against a fixed system prompt and 6 fixed examples. A change to either, or a provider-side update to the underlying model, can shift its precision-recall balance without the deploying organisation's knowledge or consent. RoBERTa-large's weights are a frozen, versioned artefact so the operating point is reproducible from the saved weights and the threshold configuration alone, independent of any third party's prompt or model version.

\textbf{Data locus.} A third-party API-based classifier sends the task description to an external provider on every call. A classifier hosted in the local network does not. Whether this matters depends on the deploying organisation's own data-handling requirements, which are outside what this paper's synthetic policy can establish.

\textbf{Verdict against the deployability standard.} No configuration swept clears both conditions of the Section 3.3 standard simultaneously; the flat 0.8 cutoff clears the precision bar (0.941) at a 27\% undershoot, the furthest from the availability bar of any RoBERTa configuration. The hybrid architecture of Section 4.2 clears the precision bar (0.940) but remains 8 percentage points short of the undershoot bar. The recommended configuration (0.8/0.6/0.4) reaches macro-F1 0.890, macro-precision 0.915, and severity-weighted delta 0.53, the best balance across the sweep, and is the configuration used in Section 6.

\section{Designing the Policy Prohibition Layer (Source 3)}
The prohibition layer enforces a set of static rules on the output of role ceilings and task-context classifier. It triggers when certain high-risk combinations of permissions exist which are not explicitly stated in the company policy as that department's workstream. When each rule triggers, it removes a defined high-risk permission which should not coexist with another, regardless of what the previous layer decided. The lethal trifecta (Willison, 2025), private data access, exposure to untrusted content, and external communication occurring together, is one textbook instance of this pattern.

The policy prohibition layer defines the rules based on the company policy and is not affected by anything within the task body. Therefore its behaviour doesn't degrade or get bypassed by adversarial inputs contrary to the task-context classifier. This makes it a good backstop layer.

\subsection{Rule Provenance and Format}
A prohibition rule states which permission to remove when a specific pair is jointly present. Each rule is scoped to exclude departments where the organisation's policy explicitly authorises the activity that would require the rule combination. Rules can either be compiled automatically from a written policy document, in the spirit of Policy-as-Prompt (Kholkar \& Ahuja, 2025), or authored by hand from the same document. This paper creates them manually and leaves automatic creation of these rules for future research. The scope is limited to evaluating the performance gains of prohibition combinations by themselves and on top of other layers in the three-source architecture, assuming they are correctly derived from the policy. The three rules below were derived from reading department-level workflows in company\_policy.txt and defining natural guardrails around them.

Rules are written in MongoDB query language (e.g. \$in, \$nin, \$all, \$gte) because it is the most widely adopted JSON-native filter syntax. The rules doesn't depend on the task definition, therefore they can't be manipulated. The full rule set, in this format, is given in Section 5.2.

Each rule definition also has a context field that contains a data sensitivity level (Public, Internal, Confidential, or Restricted, a common enterprise convention aligned with ISO/IEC 27001 Annex A) that is the threshold for the rule to trigger. These values are mapped onto the Low/Medium/High levels already used for the dataset's sensitivity metadata but they are not included in the evaluations. We decided to pin all the data access to ``Restricted'', the maximum level. It is because the prohibition layer is expected to act as a safety net when the classifier justifies the permissions but the data being accessed is outside the department responsibilities, which is the worst case scenario. The sensitivity field is included in the rule format as the threshold when the rule applies which is always true for the sake of our experiments. This assumption applies only to the prohibition-rule evaluation in Sections 5 and 6 and does not affect Section 4.

\subsection{The Rule Set}
Three rules, each tied to an explicit clause in company\_policy.txt are as follows:

\begin{lstlisting}
SOURCE_3_RULES = [
    {
        "id": "database_http",
        "when": {"permissions": {"$all": ["database_read", "http_request"]}},
        "department": {"$nin": ["Data and Analytics"]},
        "context": {"min_sensitivity": {"$gte": Sensitivity.CONFIDENTIAL}},
        "action": {"$remove": ["http_request"]},
    },
    {
        "id": "database_email",
        "when": {"permissions": {"$all": ["database_read", "email_send_external"]}},
        "department": {"$nin": ["Security", "Customer Success", "Finance", "Data and Analytics"]},
        "context": {"min_sensitivity": {"$gte": Sensitivity.CONFIDENTIAL}},
        "action": {"$remove": ["email_send_external"]},
    },
    {
        "id": "upload_http",
        "when": {"permissions": {"$all": ["file_read_uploaded", "http_request"]}},
        "department": {"$nin": []},
        "context": {"min_sensitivity": {"$gte": Sensitivity.CONFIDENTIAL}},
        "action": {"$remove": ["http_request"]},
    },
]
\end{lstlisting}
\textbf{Database-HTTP.} Removes http\_request when database\_read is also present. The policy states that analysts ``occasionally fetch live external data... via HTTP to enrich internal reports,'' which requires database\_read and http\_request combination prohibited by this rule. That's why an exemption exists for the Data and Analytics department so the rule does not block that workflow.

\textbf{Database-email.} Removes email\_send\_external when database\_read is also present, except for Security, Customer Success, Finance, and Data and Analytics. The first three departments' exemptions follow from the policy's descriptions of compliance correspondence with external auditors, renewal proposals sent to customers, and auditor and vendor communication respectively. The fourth follows from the same document's statement that analysts ``prepare reports for external stakeholders such as investors or enterprise customers, sent via email.''

\textbf{Upload-HTTP.} Removes http\_request when file\_read\_uploaded is also present. This targets the untrusted-content leg of the lethal trifecta with no department exemptions. A task reading a sensitive document and capable of making HTTP requests can use the network connection to exfiltrate or mistakenly leak the confidential data. Reading an uploaded file and outbound HTTP call is not required in combination by any department based on the policy document and the generated task dataset.

Every exemption above is created based on the responsibilities of departments referenced in the company policy document.

Section 6 evaluates this layer in combination with the other components of the proposed architecture.

\section{End-to-End Evaluation of the Architecture}
This section evaluates the assembled architecture. It reports the full system's elimination results, and then ablates the layers ahead of the prohibitions to measure what the layer contributes when each is present, degraded, or absent.

\subsection{Experimental Conditions}
Two baselines and three mechanism conditions are compared, all evaluated on the 100-record test split.

\begin{itemize}
  \item \textbf{C0}: the reference baseline. All fifteen permissions granted to every task unconditionally. This is the static all-permission provisioning this paper's architecture is designed to replace. Its severity-weighted mass is 31 per record, constant, since every record receives the same fifteen permissions.
  \item \textbf{C1}: Source 1 alone. The department role ceiling, fixed per department, granted regardless of task. This is the realistic role-based baseline.
  \item \textbf{S2}: Source 2 alone. RoBERTa-large at the recommended configuration, with no role ceiling applied. Labelled outside the C-series because it is a single source in isolation, not a step in the deployment ladder.
  \item \textbf{C2}: Source 1 $\cap$ Source 2. The intersection of the role ceiling and the classifier's output.
  \item \textbf{C3}: C2 further filtered by the prohibition layer (Source 3), using the rule set defined in Section 5.2.
\end{itemize}

\subsection{Attack-Surface Elimination}
Table 3 gives elimination by condition.

\begin{table}[t]
\caption{Attack-surface elimination by condition, against the C0 baseline.}
\label{tab:3}\centering\footnotesize\setlength{\tabcolsep}{5pt}
\begin{tabular}{lccc}
\toprule
\textbf{Condition} & \textbf{Mean raw} & \textbf{Mean sev.-weighted} & \textbf{\% of C0's} \\
 & \textbf{elimination} & \textbf{elimination} & \textbf{surface} \\
\midrule
C1 & 3.55 & 8.65 & 27.9\% \\
S2 & 12.59 & 26.11 & 84.2\% \\
C2 & 12.61 & 26.15 & 84.4\% \\
C3 & 12.61 & 26.15 & 84.4\% \\
\bottomrule
\end{tabular}
\end{table}

The full system eliminates 84.4\% of the severity-weighted attack surface exposed by static all-permission provisioning. The reduction is provided almost entirely by the task classifier which eliminates 84.2\%. The role ceiling accounts for 27.9\% when isolated, and adding it to the classifier provides only 0.2 more. This comes from the two records where the ceiling strips a permission the classifier granted. On both records the stripped permission is actually required by the task but blocked by the department ceiling (the test split's two residual ceiling violations), so the increment is bought with two additional under-grants rather than with removed over-grants. This is a direct consequence of elimination counting everything removed from the baseline (Section 3.2) regardless of usefulness. Compared to C1 rather than C0, the classifier removes 17.5 severity-weighted units per record of the 22.35 that survive the role ceiling, closing 78.3\% of the surface a role-scoped deployment still carries. The prohibition layer adds nothing measurable on top of the intersection, its scoped rules never fire on this classifier's output (Section 6.3), therefore C3 equals C2. A contribution of exactly zero invites the question of whether the prohibition layer is worth including at all. This result actually shows the deployed classifier is precise enough that no forbidden combination reaches the rules, not that the rules can never fire. As explained in Section 6.3, behind a weaker classifier those rules become operational. Section 6.4 measures the benefit of the layer when the classifier it backs up is removed, and Section 7 draws the methodological consequence for evaluating layered defences.

\subsection{Rule Scoping and Firing Behaviour Behind the Classifier}
Section 5.2 states as a design requirement that no rule in the final set blocks a combination the policy authorises. This section evaluates the scoped rule set against C2 and a deliberately unscoped version of the same prohibitions to show why the department exemptions matter. It reverts each prohibition to its literal form, under which database\_read may not co-occur with email\_send\_external or http\_request in any department. Table 4 compares the scoped and unscoped rule sets.

\begin{table}[t]
\caption{The prohibition rules applied to C2.}
\label{tab:4}\centering\footnotesize\setlength{\tabcolsep}{4pt}
\begin{tabular}{lccc}
\toprule
\textbf{Condition} & \textbf{Sev.-weighted} & \textbf{Under-} & \textbf{Newly unsat.} \\
 & \textbf{elim. (\% of C0)} & \textbf{shoot} & \textbf{vs.\ C2} \\
\midrule
C2 & 84.4\% & 23\% & --- \\
C3, unscoped & 85.0\% & 28\% & 5 records \\
C3, scoped (Section 5.2) & 84.4\% & 23\% & 0 records \\
\bottomrule
\end{tabular}
\end{table}

The scoped rule set introduces no new failures relative to C2. It never removes a permission that is required on a record C2 had already satisfied. This holds for all six RoBERTa-large threshold configurations, from the loosest (0.5/0.3/0.1, 41\% overshoot) to the tightest. With Claude Haiku, three of the six configurations fire. At flat 0.5 and at 0.6/0.4/0.2 a single firing each correctly removes a genuine over-grant of a prohibited pair, and at 0.5/0.3/0.1 the rules fire on four records, three of which lose a permission the task needed. This finding establishes that the rules do begin to engage once a classifier's overshoot becomes extreme (Haiku's loosest configuration carries a severity-weighted delta of 2.88, more than five times the deployed classifier's), so the rules not firing is a property of a sufficiently precise classifier, not of the rules being unreachable.

Evaluated against C2, the unscoped version fires on 7 of 100 records. For six of seven firings, the classifier predicted correctly in each case but the unscoped rule removed a correct prediction. The other one is a correct removal of a genuine classifier over-grant. From six incorrect removals, one case already had unmet ground truth in C2 for an unrelated reason, so the rule's removal does not change that record's status.

An unscoped rule does not distinguish if a workflow is authorised in the policy. There are four Data and Analytics firings that have database\_read/http\_request combination the policy explicitly authorises for that department (Section 5.2). The other three firings are the database\_read/email\_send\_external combination the policy explicitly authorises for Finance. This is the evidence that motivates scoping each rule to the department exemptions in Section 5.2, and the rule set evaluated everywhere else in this paper is the scoped version.

This result validates the rule set itself is sound, and doesn't demonstrate performance of the rules. The rules were derived by the author reading company\_policy.txt directly. The dataset's ground-truth labels were generated independently by an LLM reading the same document during dataset construction. Therefore, the rules having zero availability impact shows the agreement of two readings and is the expected result of correct rule derivation. The swept TF-IDF baseline shows the same pattern as Claude Haiku. The rules never fire at its four tighter configurations and fire only at the two loosest, three times at 0.6/0.4/0.2 and eight times at 0.5/0.3/0.1, in every case on the Upload-HTTP pair.

\subsection{The Policy Prohibition Layer in Ceiling Fallback}
Section 6.3 tests the policy prohibition layer with the classifier present in some form. This section tests the case where the classifier is absent or bypassed entirely, so the system falls back to whatever the role ceiling alone would grant. The following question is answered: If an attacker successfully bypassed the classifier, how much risk do prohibitions still prevent, and does it cost anything on tasks with no attacker involved at all?

The first half of that question is department-level and task-independent by construction. Once the classifier is out of the picture, the prohibition layer evaluates a department's role ceiling, the same set for every task in that department, not any individual task, so its effect can only be measured per department, not per task. Table 5 gives the per-department ceiling reduction.

\begin{table}[t]
\caption{Severity-weighted ceiling reduction by the prohibition layer when the classifier is absent.}
\label{tab:5}\centering\footnotesize\setlength{\tabcolsep}{5pt}
\begin{tabular}{lccc}
\toprule
\textbf{Department} & \textbf{Ceiling} & \textbf{After} & \textbf{Reduction} \\
 & \textbf{(sev.-wtd)} & \textbf{prohibitions} & \\
\midrule
Engineering & 26 & 23 & 11.5\% \\
Data and Analytics & 22 & 19 & 13.6\% \\
Security & 26 & 23 & 11.5\% \\
Customer Success & 21 & 21 & 0\% \\
Finance & 19 & 19 & 0\% \\
Legal and Compliance & 16 & 16 & 0\% \\
\bottomrule
\end{tabular}
\end{table}

Three departments don't show any reduction because they don't have a forbidden permission pair within their role ceiling. The other three show reductions between 11.5\% and 13.6\%.

Risk reduction metric is measured against all 600 records since data split is not required for the prohibition layer. A task is counted as newly blocked if its ground truth was satisfiable by the role ceiling alone and is not satisfiable once the prohibition rules are applied. Table 6 gives the block rates.

\begin{table}[t]
\caption{Legitimate tasks newly blocked by the prohibitions in ceiling fallback.}
\label{tab:6}\centering\footnotesize\setlength{\tabcolsep}{4pt}
\begin{tabular}{lcccl}
\toprule
\textbf{Department} & \textbf{Tasks} & \textbf{Blocked} & \textbf{Rate} & \textbf{Rule} \\
\midrule
Engineering & 150 & 9 & 6.0\% & Database-HTTP \\
Data and Analytics & 90 & 28 & 31.1\% & Upload-HTTP \\
Security & 90 & 20 & 22.2\% & Database-HTTP \\
Customer Success & 120 & 0 & 0\% & --- \\
Finance & 90 & 0 & 0\% & --- \\
Legal and Compliance & 60 & 0 & 0\% & --- \\
\bottomrule
\end{tabular}%
\par\vspace{5pt}%
\parbox{\linewidth}{\scriptsize Five of the 600 records are excluded from
metrics as they have ground truth outside their department ceiling in the
first place.}%
\end{table}

Every department with a nonzero reduction also has a nonzero block rate. Data and Analytics has both the largest reduction (13.6\%) and the highest cost (31.1\%), because its ordinary workload most closely resembles the pattern the Upload-HTTP and Database-HTTP rules are built to catch. The reader is urged to notice the Data and Analytics friction here is from Upload-HTTP, unrelated to the unscoped-rule problem shown earlier (Section 6.3). Every one of the 28 blocked Data and Analytics tasks need http\_request without needing file\_read\_uploaded, the two are never jointly required anywhere in the dataset (Section 5.2), but file\_read\_uploaded sits in every department's role ceiling as a baseline capability, so the Upload-HTTP rule fires regardless of what the task itself is doing, removing the capability of the agent making HTTP requests. This causes the tasks to fail. The Database-HTTP rule produces the identical pattern for the Security and Engineering departments whose role ceilings hold database\_read and http\_request. 19 of the 20 blocked Security tasks require http\_request permission without needing database\_read. However all 20 of them lose it because database\_read happens to sit in the department role ceiling and the rule strips http\_request when triggered. These results show policy ceilings are not feasible to apply statically in task initialization, discussed in detail in the next section.

The Database-email rule contributes no blocked tasks in either table because no department excluded from its exemption list holds database\_read and email\_send\_external together in its role ceiling.

\subsection{Static Versus Runtime Enforcement of the Prohibitions}
Section 6.4's results show that removing credentials preemptively can cause legitimate permissions to be pruned. A static implementation, the one evaluated throughout this paper, checks a complete candidate permission set before a task begins and removes one member of any prohibited pair found present. Applied to a set the classifier has already narrowed to task-relevant permissions, this correctly distinguishes a task using both halves of a pair from a task using one. However, applied to a department's full role ceiling, it cannot distinguish whether a task intends to use that permission, because the ceiling holds both halves whenever the department is permitted both. Section 6.4's entire cost column assumes this static implementation, which is proven to be an infeasible permission stripping method.

A runtime implementation would instead monitor the sequence of permissions actually invoked during execution. If a task invokes the first permission of a prohibited pair and later attempts to invoke the second, the attempt is blocked (enforcing deployment) or logged (observe-only deployment) at the point of the second invocation. A task that doesn't use a permission in its department ceiling doesn't trigger any rules. This removes Section 6.4's cost in principle since 56 of the 57 blocked tasks in that section don't invoke both halves of the pair that blocks them.

This paper does not evaluate the runtime implementation. It would require monitoring the order of invocation of the permissions in the dataset during task execution. Evaluating it requires a live agent capable of taking actions in sequence against a controlled environment. ControlArena side-task evaluations are a natural setting for this once such an agent exists, and building it is noted as future work.

The static and runtime modes are two different deployment states useful in different conditions. Static deployment preemptively sets the permissions, and abstains from deploying a credential if it is not expected to be used by the task. This effectively removes the attack surface, and makes credential exfiltration infeasible. Runtime enforcement allows the prohibition rules to be applied dynamically when the task actually attempts to use them. It creates a safety net when the classifier is degraded, bypassed, or simply absent from a given deployment, while not preemptively removing credentials, preventing the failure modes of Section 6.4. The tradeoff is that it creates a system-level enforcement layer which is bypassable in theory by privilege escalation exploits or side channel attacks. Therefore the ideal implementation of such a design would be static deployment of the role ceiling and the classifier, and dynamic deployment of the prohibitions.

\section{Discussion and Limitations}
\textbf{The residual error is due to insufficient taxonomy.} Three independent results show that the quality of the classifier is not the limiting factor for getting better results. First, the classifier's remaining over-granting concentrates on a short list of permissions with database\_read alone carrying 34\% of the severity-weighted residual overshoot (Section 4.2). Second, the most efficient improvement in the entire classifier evaluation came from a change to the decision structure: the hybrid of Section 4.2 was the only intervention that improved precision, availability, and F1 simultaneously. Every other route to that precision gives something up. Tightening the threshold to a flat 0.8 reaches similar precision but raises undershoot from 21\% to 27\%. Replacing RoBERTa-base with RoBERTa-large buys its precision gain with three times the parameters (Section 4.1). Once one permission's decision was handed to a rule, precision cleared the deployability bar the classifier alone never reached. Third, the entire ceiling-fallback friction cost of Section 6.4 is caused by database\_read and file\_read\_uploaded. These permissions are too coarse to separate a safe use from a dangerous one. The risk of data access in both permissions depends on the sensitivity of the data being accessed. Permission database\_read is the common element of the first and third results, which makes it the single highest-value target for refinement.

Above roughly 0.89 macro-precision, further investment in the classifier is not beneficial, because the residual errors are not classification mistakes the model could correct with more capacity or better thresholds. In most cases, a single permission label conflates two actions with different risk levels which is not solvable by any classifier due to lack of granularity. The productive next step is to refine the taxonomy so that the distinctions the rules need to make are expressible as separate permissions, for example splitting database\_read into sensitivity-qualified variants (Section 8). This reframes the architecture's development priority from classifier to the permission ontology.

\textbf{Evaluating backstop layers.} Measuring the prohibition rules on normal traffic with every layer active, the layer doesn't contribute anything (Section 6.2). This result alone would justify removing it. However when it is measured with previous layers ablated, it removes 11.5-13.6\% of three departments' ceilings (Section 6.4). This shows that the contribution of a backstop layer is invisible in aggregate performance and appears only when the other layers fail. This would happen in a real world setting due to operational failures or adversarial prompts. Evaluating a defence-in-depth component by its marginal effect on end-to-end metrics therefore systematically undervalues the security value of it. The correct evaluation method is to ablate each preceding layer and measure the component against the resulting exposure like described in Section 6.1.

\textbf{Scalable oversight.} The task-context classifier is a weaker and cheaper model supervising a strong, expensive one, and Section 4 shows the capability gap between them does not degrade the supervision. A 355M-parameter encoder fine-tuned in approximately fourteen minutes on an NVIDIA T4 GPU reaches the classification quality of a frontier model on the permission-prediction task (Section 4.2). AI control protocols generally operate under the constraint that the trusted model is less capable than the untrusted model it oversees (Greenblatt et al., 2024). Permission prediction avoids this gap because it is a narrow multi-label classification over the user's task description instead of relying on open-ended evaluation of the agent's behaviour or reasoning. The supervised agent can grow more capable without the classifier needing to grow with it, because the task scope is not tied to the inference in any way. The scalable-oversight margin for this specific control task is therefore wide. This does not generalise to control tasks that require judging the agent's outputs rather than the user's inputs, where the capability gap does bind.

The permission gate classifier is cheap enough to be fine-tuned based on the company policy and retrained when the policy evolves.

\textbf{The price of role granularity.} Least privilege at the granularity of individual tasks has been the stated goal since Saltzer \& Schroeder (1975), and enterprise access control settled on role granularity instead, because re-provisioning a human principal per task costs human time and coordination. Section 6.2 quantifies what that settlement costs in attack surface showing that the role ceiling alone closes 27.9\% of the severity-weighted surface static provisioning creates, while adding task-context scoping closes 84.4\%. The 3x gap between the two is the measured price of role-granular rather than task-granular provisioning. An agent's task arrives as machine-readable text, so the scoping decision is a classifier inference rather than a ticket queue, and the per-task cost that made task granularity impractical for human principals falls to milliseconds. In this one dimension an agent is more governable than a human employee since its unit of work is legible to an automated policy at the moment the work is assigned.

\textbf{Implications for agent provisioning.} Failure modes of deployments that grant each agent host a static, standing credential set are documented in security evaluations of OpenClaw-style deployments (Wang et al., 2026). This type of deployment ties a set of credentials to a host identity for the lifetime of that host, whether or not the host is actively working on a task that needs them. When a task-scoped classifier is used the permission set is specifically computed from the task definition and is not tied to a host or user's identity. This means an agent instance no longer needs to pre-exist with a standing credential set. The agent can be allocated for the duration of a single task with exactly the required permissions/credentials deployed, and torn down once the task completes or times out. This method is closer to a serverless function invocation (Jonas et al., 2019) such as AWS Lambda than to a persistently provisioned host. The security architecture evaluated in this paper is a precondition for that allocation model since it allows dynamic handling of credential sets based on the task and principal together.

\textbf{Limitations.} Every friction figure in Section 6.4 inherits the limitations of the synthetic dataset that were documented at release (Noyan, 2026, §6). Classifier metrics are measured on a 100-record test split which makes the 0.5 points macro-F1 gap between Claude Haiku and RoBERTa-large negligible. For that reason Section 4 focuses on precision metrics and deployment properties of two classifiers rather than the F1 measurements.

The rule set used in the prohibition layer was derived by a single author, and the fidelity of deriving rules from policy text (by human or by LLM) is not measured (Section 5.1). The data sensitivity gate is evaluated only at its worst-case constant (Section 5.1), so Section 6.4's friction rates are only upper bounds under that assumption.

The runtime enforcement mode that Section 6.5 identifies as the correct standalone deployment of the prohibitions is designed but not evaluated, for lack of execution-trace data. The multi-step scope-expansion limitation of the underlying architecture carries over from the prior work unchanged.

\section{Future Work}
Every prompt in the dataset describes a legitimate task, so all three layers have been measured on benign traffic only. A set of adversarial task descriptions crafted to cause over-grants through prompt injection and context manipulation can test the layers under attack. Section 6.3 shows the policy derived rules become operational when the Source 2 classifier is too permissive, therefore adversarial traffic can highlight the performance of policy prohibitions using crafted tasks targeting overshoot.

A future iteration will test the architecture against adversarial input using existing AI control benchmarks. The runtime enforcement mode of Section 6.5 requires an agent-side monitor over system calls and an execution environment to evaluate them. A system-level agent-side monitor can be implemented and run in ControlArena side-task evaluations (UK AI Security Institute \& Redwood Research, 2026) to test it directly. Its results can be compared against existing propensity and deception benchmarks to measure the improvement the three-layer architecture provides. The same architecture can also be tested against model organisms of emergent misalignment (Turner et al., 2025).

Another benchmarking improvement includes the compute usage evaluation of classifiers. Section 4.3 makes a compute-location argument but doesn't run the benchmarks to get a quantitative footing. A controlled latency and cost benchmark can measure per-request figures for RoBERTa-large on stated on-host hardware and for Claude Haiku 4.5 over the API, at the dataset's actual prompt-length distribution.

The synthetic dataset is the biggest limitation of this paper and is the strongest candidate for improvement. Section 6.4 evaluates the sensitivity gate only at its constant worst case, because the dataset's Low/Medium/High labels are too unreliable to use (Noyan, 2026, §5.3). Re-labelling the tasks that require database\_read or file\_read\_uploaded under an explicit PII criterion would strengthen the dataset and allow more granular rules. As explained in Section 7, the most immediate improvement point is the permission granularity rather than classifier accuracy. The dataset improvements will enable testing the rules with sensitivity-qualified permission variants (database\_read:low\_sensitivity, database\_read:high\_sensitivity).

Classifier tuning is an alternative forward path for this research. The classifier thresholds can be calibrated per permission risk tier (currently static e.g. low: 0.3, medium: 0.5, high: 0.7). Moreover, the classifier both trains and evaluates on the same six departments, so the results cannot separate learning the policy's structure from memorising per-department priors. Holding a single test department out for training and evaluating that department can show if the classifier is resilient to onboarding a new organisational unit without re-training.

Automated compilation of the rules from policy text has been deliberately out of scope (Section 5.1) and the rules are manually generated. Rule generation from policy needs its own evaluation for derivation fidelity, in the way this paper evaluates the classifier.

\section{Conclusion}
Least privilege at the granularity of individual tasks has been the stated ideal of access control since Saltzer and Schroeder (1975), and enterprises settled for role granularity because no one could re-provision a human principal for every task. The evaluation shows that for AI agents this settlement is no longer forced. An agent's task arrives as machine-readable text, so the permission set can be computed at the moment work is assigned via role, policy and task-context enforcements together. A credential withheld at provisioning won't exist in the system and therefore cannot be misused. This is independent from the agent's reasoning or evasion sophistication, so the claims are expected to hold as the agents grow more capable. The results show that task-granular least privilege for agents is feasible, and the research outlines what stands between feasibility and deployment.

\ifanon\else
\section*{Acknowledgements}
This work grew out of discussions during the BlueDot Impact Technical
AI Safety Project course.
\fi

\section*{Open Science}
\ifanon
\else
The dataset, company policy document, training and evaluation code, cached probability matrices, and the experiment results are available at github.com/0xballistics/mostargate. The fine-tuned RoBERTa-large checkpoint is published at huggingface.co/buraknoyan/mostargate-c2-roberta-large.
\fi

\section*{LLM Usage Considerations}
All results, tables and figures in this paper were produced by the author running the experiments. The content of the paper was written by the author, then modified and corrected using LLM assistance. All LLM edits in this paper are vetted and validated. Code for reproducing the results is created with assistance of LLM coding tools (Claude variants) but the code itself and the output is verified by the researcher. Likewise, the dataset is synthetically generated by an LLM and validated against independent expert re-annotation. The artifacts and source code is shared in the public repository. LLMs are also used as subjects of experiment as explained throughout the paper (Claude Haiku).

\section*{References}
\begin{list}{}{\leftmargin=1em \itemindent=-1em \itemsep=2pt \parsep=0pt \topsep=4pt}
\item Amazon Web Services. ``What is IAM?''. \url{https://docs.aws.amazon.com/IAM/latest/UserGuide/introduction.html}, 2024. AWS Identity and Access Management User Guide.
\item Cohen, J. A coefficient of agreement for nominal scales. Educational and Psychological Measurement, 20(1):37--46, 1960.
\item Dennis, J. B. and Van Horn, E. C. Programming semantics for multiprogrammed computations. Communications of the ACM, 9(3), 1966.
\item Greenblatt, R., Denison, C., Wright, B., Roger, F., MacDiarmid, M., Marks, S., Treutlein, J., Belonax, T., Chen, J., Duvenaud, D., Khan, A., Michael, J., Mindermann, S., Perez, E., Petrini, L., Uesato, J., Kaplan, J., Shlegeris, B., Bowman, S. R., and Hubinger, E. Alignment faking in large language models. arXiv preprint arXiv:2412.14093, 2024.
\item Greshake, K., Abdelnabi, S., Mishra, S., Endres, C., Holz, T., and Fritz, M. Not what you've signed up for: Compromising real-world LLM-integrated applications with indirect prompt injection. In AISec '23 (Proceedings of the 16th ACM Workshop on Artificial Intelligence and Security), pp. 79--90, 2023. doi: 10.1145/3605764.3623985.
\item Hubinger, E., Denison, C., Mu, J., Lambert, M., Tong, M., MacDiarmid, M., Lanham, T., Ziegler, D. M., Maxwell, T., Cheng, N., Jermyn, A., Askell, A., Radhakrishnan, A., Anil, C., Duvenaud, D., Ganguli, D., Barez, F., Clark, J., Ndousse, K., Sachan, K., Sellitto, M., Sharma, M., DasSarma, N., Grosse, R., and Kravec, S. Sleeper agents: Training deceptive llms that persist through safety training. arXiv, 2024. doi: 10.48550/arxiv.2401.05566.
\item Inan, H., Upasani, K., Chi, J., Rungta, R., Iyer, K., Mao, Y., Tontchev, M., Hu, Q., Fuller, B., Testuggine, D., and Khabsa, M. Llama guard: Llm-based input-output safeguard for human-ai conversations, 2023. URL \url{https://arxiv.org/abs/2312.06674.}
\item Jonas, E., Schleier-Smith, J., Sreekanti, V., Tsai, C.-C., Khandelwal, A., Pu, Q., Shankar, V., Carreira, J., Krauth, K., Yadwadkar, N., Gonzalez, J. E., Popa, R. A., Stoica, I., and Patterson, D. A. Cloud programming simplified: A Berkeley view on serverless computing. arXiv preprint arXiv:1902.03383, 2019.
\item Kholkar, G. and Ahuja, R. Policy-as-prompt: Turning ai governance rules into guardrails for ai agents, 2025.
\item Meinke, A., Schoen, B., Scheurer, J., Balesni, M., Shah, R., and Hobbhahn, M. Frontier models are capable of in-context scheming. arXiv preprint arXiv:2412.04984, 2024. Apollo Research.
\item Miller, M. S. Robust Composition: Towards a Unified Approach to Access Control and Concurrency Control. PhD thesis, 2006.
\ifanon
\else
\item Noyan, H. B. Dynamic capability scoping for enterprise AI agents: A synthetic dataset and three-source permission architecture. In *Second Workshop on Agents in the Wild: Safety, Security, and Beyond (AIWILD) at ICML*, 2026. URL \url{https://openreview.net/pdf?id=fTdw60ITBv}
\fi
\item OWASP Foundation. OWASP top 10 for LLM applications 2025. \url{https://genai.owasp.org/llm-top-10/}, 2025.
\item Palumbo, N., Choudhary, S., Choi, J., Chalasani, P., and Jha, S. Policy compiler for secure agentic systems, 2026.
\item Rebedea, T., Dinu, R., Sreedhar, M. N., Parisien, C., and Cohen, J. Nemo guardrails: A toolkit for controllable and safe llm applications with programmable rails. In Proceedings of the 2023 Conference on Empirical Methods in Natural Language Processing: System Demonstrations, pp. 431--445. Association for Computational Linguistics, 2023. doi: 10.18653/v1/2023.emnlp-demo.40. URL \url{https://aclanthology.org/2023.emnlp-demo.40.}
\item Rose, S., Borchert, O., Mitchell, S., and Connelly, S. Zero trust architecture. Technical Report NIST SP 800-207, National Institute of Standards and Technology, 2020.
\item Saltzer, J. H. and Schroeder, M. D. The protection of information in computer systems. Proceedings of the IEEE, 63(9):1278--1308, 1975.
\item Shapiro, J. S., Smith, J. M., and Farber, D. J. EROS: A fast capability system. In Proceedings of the 17th ACM Symposium on Operating Systems Principles (SOSP), 1999.
\item Tsai, L. and Bagdasarian, E. Contextual agent security: A policy for every purpose. In Proceedings of the Workshop on Hot Topics in Operating Systems (HotOS '25), 2025. doi: 10.1145/3713082.3730378.
\item Turner, E., Soligo, A., Taylor, M., Rajamanoharan, S., and Nanda, N. Model organisms for emergent misalignment. arXiv preprint arXiv:2506.11613, 2025.
\item UK AI Security Institute and Redwood Research. Controlarena: A library for running AI control experiments. \url{https://github.com/UKGovernmentBEIS/control-arena}, 2026.
\item Wang, Y., Gao, H., Niu, Z., Liu, Z., Zhang, W., Wang, X., and Lian, S. A systematic security evaluation of openclaw and its variants. arXiv preprint arXiv:2604.03131, 2026.
\item Willison, S. The lethal trifecta for ai agents: Private data, untrusted content, and external communication. \url{https://simonwillison.net/2025/Jun/16/the-lethal-trifecta/}, June 2025. Blog post.
\end{list}
\end{document}